\documentclass[letterpaper]{article} 
\usepackage[]{aaai23}  
\usepackage{times}  
\usepackage{helvet}  
\usepackage{courier}  
\usepackage[hyphens]{url}  
\usepackage{graphicx} 
\def\UrlFont{\rm}  
\usepackage{natbib}  
\usepackage{caption} 
\newcommand{\pr}[1]{\textsf{#1}}

\newcommand{\PO}[0]{\pr{PO}}
\newcommand{\TPP}[0]{\pr{TPP}}

\newcommand{\DC}[0]{\pr{DC}}

\newcommand{\EQ}[0]{\pr{EQ}}
\newcommand{\EC}[0]{\pr{EC}}
\newcommand{\NTPP}[0]{\pr{NTPP}}
\newcommand{\NTPPi}[0]{\pr{NTPPi}}
\newcommand{\TPPi}[0]{\pr{TPPi}}

\newcommand{\chat}[0]{ChatGPT-4}

\newcounter{expnumcounter}
\def\expnum{\addtocounter{expnumcounter}{1}\theexpnumcounter}

\usepackage{algorithm}
\usepackage{algorithmic}

\usepackage{newfloat}
\usepackage{listings}
\DeclareCaptionStyle{ruled}{labelfont=normalfont,labelsep=colon,strut=off} 
\floatstyle{ruled}
\newfloat{listing}{tb}{lst}{}
\floatname{listing}{Listing}
\title{An Evaluation of \chat's \\ Qualitative Spatial Reasoning Capabilities in RCC-8 }
\author{
    A G Cohn
}
\affiliations{
    University of Leeds \& The Alan Turing institute, UK\\

}

\usepackage{bibentry}

\begin{document}

\maketitle

\begin{abstract}
Qualitative Spatial Reasoning (QSR) is well explored area of Commonsense Reasoning and has multiple applications ranging from Geographical Information Systems to Robotics and Computer Vision.  Recently many claims have been made for the capabilities of Large Language Models (LLMs). In this paper we investigate the extent to which one particular LLM can perform classical qualitative spatial reasoning tasks on the mereotopological calculus, RCC-8.
\end{abstract}

\section{Introduction }
\begin{figure}[htb]
\centering
\includegraphics[width=0.99\columnwidth]{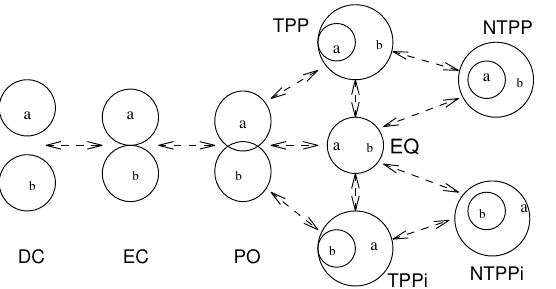} 
\caption{The eight relations of the RCC-8 calculus illustrated in 2D.}
\label{rcc8-cn-diagram}
\end{figure}

Qualitative Spatial Reasoning (QSR\footnote{We may use QSR as shorthand for both Qualitative Spatial Reasoning and Qualitative Spatial Representation; context should usually make clear which is intended.}) \cite{cohn2008qualitative,chen2015survey,cohn2001qualitative} is a well developed field which is concerned with the representation of qualitative spatial information and reasoning with it.  In natural language, spatial information is usually represented qualitatively  (using prepositions such as \emph{on, in, left of, part of, under, touching, ...)} and many calculi have been developed to represent such information.  There are calculi for mereological relations (such as RCC-5 \cite{DBLP:journals/jair/JonssonD97}), mereotopological relations (such as RCC-8 \cite{randell1992spatial,cohn1997qualitative}), directions (such as OPRA \cite{moratz2006representing}), size \cite{gerevini2002combining} for example as well as calculi combining two different aspects of spatial information, such as the Rectangle Algebra \cite{guesgen1989spatial,DBLP:conf/aaai/MukerjeeJ90} which can represent both mereotopological information as well as directional.  What is common to all these calculi is that they consist of a set of \emph{jointly exhaustive and pairwise disjoint} (JEPD) \emph{base} relations.  For example, RCC-8 contains eight JEPD \emph{base} relations, illustrated in 2D in Fig. \ref{rcc8-cn-diagram}.

\emph{Large Language Models} (LLMs) \cite{devlin-etal-2019-bert,brown2020language}, such as \chat\  \cite{roumeliotis2023chatgpt} are a recent example of so called \emph{Foundation Models} which have been trained on very large textual corpora in order to generate text in response to a prompt. This is not the place to survey this burgeoning field, but we note that many claims have been made for the power and apparent intelligent behaviour that  these models can display.  In particular their performance on some benchmarks may lead one to believe that they possess, at least to some degree, the ability to perform commonsense reasoning.  Spatial reasoning is usually regarded as one core aspect of common sense so it is natural to ask whether LLMs can reason about qualitative spatial information. This is the question  that we address here.

In earlier work \cite{cohn2023dialectical} we use extended dialogues with an LLM to try to map the boundaries of spatial commonsense in some LLMs, addressing a variety spatial challenges, and examining not only the response given but also the explanation/justification of the response, but did not specifically focus on existing QSRs, though some questions were asked which do correspond to particular reasoning steps in an existing QSR.  Here we focus on one specific QSR and ask the question as to what extent an LLM can perform reasoning in that calculus, and conduct a more exhaustive evaluation, but looking at the ability to perform compositions between relations and also to reason about the conceptual neighbourhood diagram of the calculus.  Weaknesses in the reasoning powers of LLMs have previously been noted (e.g. \cite{cai2023human}) so one might not expect LLMs to perform well in this regard. But on the other hand, there are a large number of papers about QSR in the literature and these are likely to have formed part of the  training corpus of an LLM, and thus might facilitate correctly responding to prompts --  though the information concerning the actual reasoning steps are often given in tables (in particular \emph{composition tables} -- see below) and thus might be hard for LLM training procedures to process well.

There are now many LLMs in the literature.  Some of these are open source and are explicit about the training corpus; others are closed and give no specific information about the training, or the precise corpus, such as the GPT family of LLMs.  Nevertheless since we observed previously \cite{cohn2023dialectical} that \chat\  and GPT4 were  the most performant for spatial reasoning, we use \chat\  as the LLM with which we perform our experiments. In each of the experiments below, an initial prompt gave the problem setting and the task to be performed. Subsequent prompts in the conversation probed about one specific inference (e.g. one cell in a composition table). Each experiment was a separate conversation and the version number of \chat\  was given as ``May 24'' (presumably, the version of May 24 2023). The conversations can be found at the following location: https://tinyurl.com/qr23sup.

\section{Compositional Reasoning}
\begin{figure*}[htb]
\centering
\includegraphics[width=0.99\textwidth]{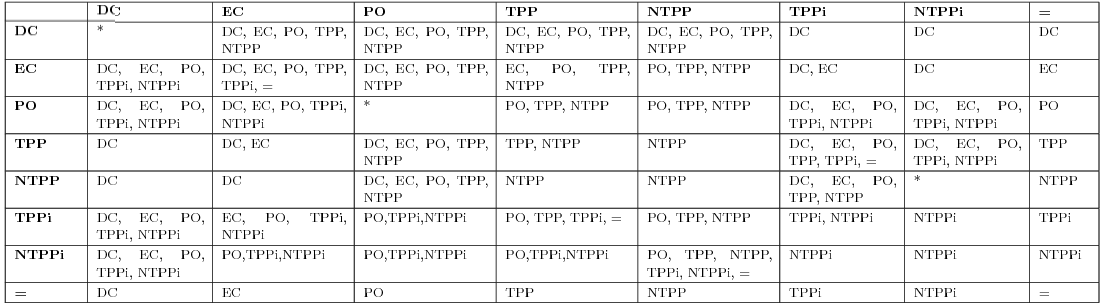} 
\caption{The RCC-8 Composition Table \cite{cohn1997qualitative}}
\label{rcc8comptable}.
\end{figure*}

The most researched form of reasoning with Qualitative Spatial Relations is that of composition: i.e. given a two facts R1($x,y$), and R2($y,z$), then what relations are possible between $x$ and $z$? In general more than one base relation is possible, so the answer may take the form of a disjunction of relations, as can be seen in the composition table for RCC-8 in Fig. \ref{rcc8comptable}. Given the ubiquity of RCC-8 in the QSR literature, in this paper we focus on the abilities of an LLM to reason with RCC-8.  

\subsection{Experiment \expnum: Compositional Reasoning in RCC-8}
In this experiment the standard names of the eight relations were given as well as a sentence stating that the relations were part of the RCC-8 calculus:
\begin{quote}
\textit{   
Consider the following set of eight pairwise disjoint and mutually exhaustive binary spatial relations. These relations form part of the well known RCC-8 qualitative spatial reasoning calculus. DC(x,y) means that x and y are disconnected and share no spatial parts. EC(x,y) means that x and y touch at a boundary but do not share any interior parts. PO(x,y) means that x and y share a spatial part, but neither is part of the other.  TPP(x,y) means that x is part of y and touches y’s boundary. NTPP(x,y) means that x is part of y but does not touch y’s boundary.  TPPi(x,y) is the same as TPP(y,x).  NTPPi(x,y) is the same as NTPP(y,x).  Finally, EQ(x,y) means that x and y are coincident. I will now ask you a series of questions about these relations. There may be more than one possible relation, in which case name all of the possible answers. If all eight relations are possible, just say ALL.}
\end{quote}

As an example of a prompt and \chat's response, consider the first one:
\begin{quote}
{\bf Prompt:} \emph{If DC(x,y) and DC(y,z) then what are the possible relationships between x and z.} 
{\bf Response: }
\textit{If DC(x,y) and DC(y,z), then x and z can either be disconnected (DC) or externally connected (EC). This is because the disconnection between x and y, and y and z, only guarantees that neither x nor z is connected to y. It does not preclude the possibility that x and z themselves might be disconnected or touching at a boundary.
However, the other relations like part-of (TPP, NTPP) and their inverses (TPPi, NTPPi), partial overlap (PO), and equality (EQ) are not possible because they would imply some level of connection between x and z through y, which contradicts our initial conditions. So, the possible relationships between x and z are:
DC(x,z): x and z are disconnected.
EC(x,z): x and z are externally connected.}
\end{quote}
Notice that in fact the composition of \DC\  with \DC\  is the union of all eight RCC-8 relations, so the reasoning is incorrect here.

\begin{figure}[htb]
\centering
\includegraphics[width=0.99\columnwidth]{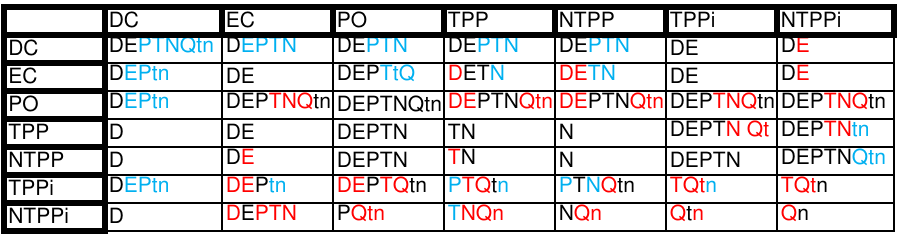} 
\caption{The Composition Table for RCC-8 produced by \chat. The entry in each cell uses the following coding: {\sf D (DC), E(EC), P(PO), T(TPP), N(NTPP), t(TPPi), n(NTPPi), Q(EQ).} Black means that relation is correctly predicted (85 times), red means that relation is incorrectly predicted (61 times), blue means that the relation was incorrectly not predicted (49 times). 
 }
\label{rcc8-composition-table}.
\end{figure}

\begin{figure}[htb]
\centering
\includegraphics[width=0.99\columnwidth]{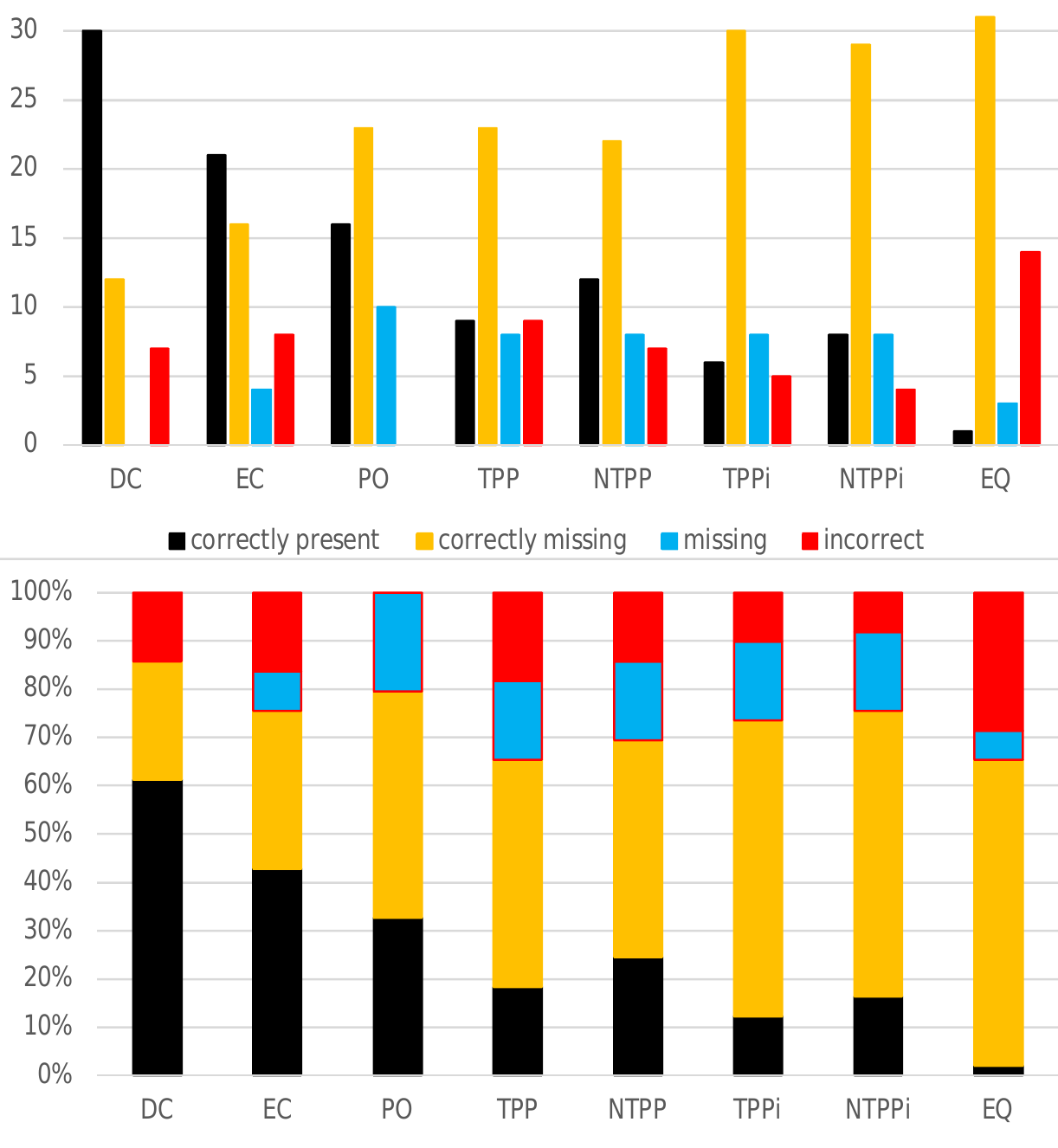} 
\caption{Relation statistics for the Composition Table for RCC-8 produced by \chat. The upper chart shows the absolute number of relations, and the lower the relative percentage for each relation. 
 }
\label{rcc8-comp-relation-stats}.
\end{figure}

The results of the 7*7=49  prompts issued (compositions involving \EQ\  were omitted since they should be trivial) are summarised in  Fig. \ref{rcc8-composition-table}. There is a total of 85 relations correctly predicted (those in black),  197 relations which are correctly not predicted (those not in the table). Thus there are a total of 282 correctly predicted (both present and missing) relations, i.e. 71.94\% of the total possible (7*7*8 = 392). 11 of the cells (out of 49) are completely correct. This is clearly much better than chance but far from a correct composition table.  Fig. \ref{rcc8-comp-relation-stats} shows the absolute number and relative percentages of the 8 relations in the proportion that they are correctly or incorrectly predicted. As can be seen, \DC\  is never missed as a correct prediction, and \PO\  is never incorrectly predicted as present. \DC\  has the highest accuracy which is perhaps not surprising since \DC\  might be regarded as intuitively easier to reason about (and this fits with the predictions about human preferred models in the section below). Including both correctly predicted and correctly missing relationships, all eight relations have broadly similar accuracies.

\begin{figure}[htb]
\centering
\includegraphics[width=0.99\columnwidth]{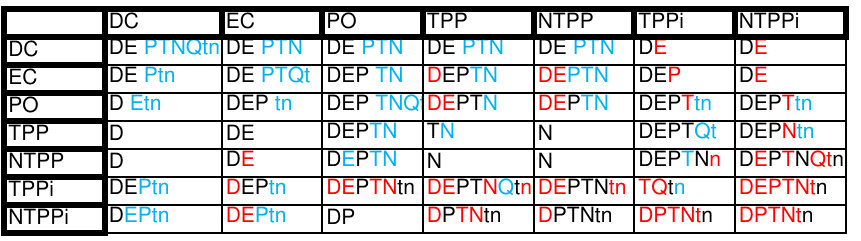} 
\caption{The Composition Table for RCC-8 produced by \chat\  using anonymised relations. The entry in each cell uses the following coding: {\sf D (DC), E (EC), P (PO), T (TPP), N (NTPP), t (TPPi), n (NTPPi), Q (EQ).} Black means that relation is correctly predicted (95 times), red means that relation is incorrectly predicted (57 times), blue means that the relation was incorrectly not predicted (72 times). 
 }
\label{xrcc8-composition-table}.
\end{figure}

\begin{figure}[htb]
\centering
\includegraphics[width=0.99\columnwidth]{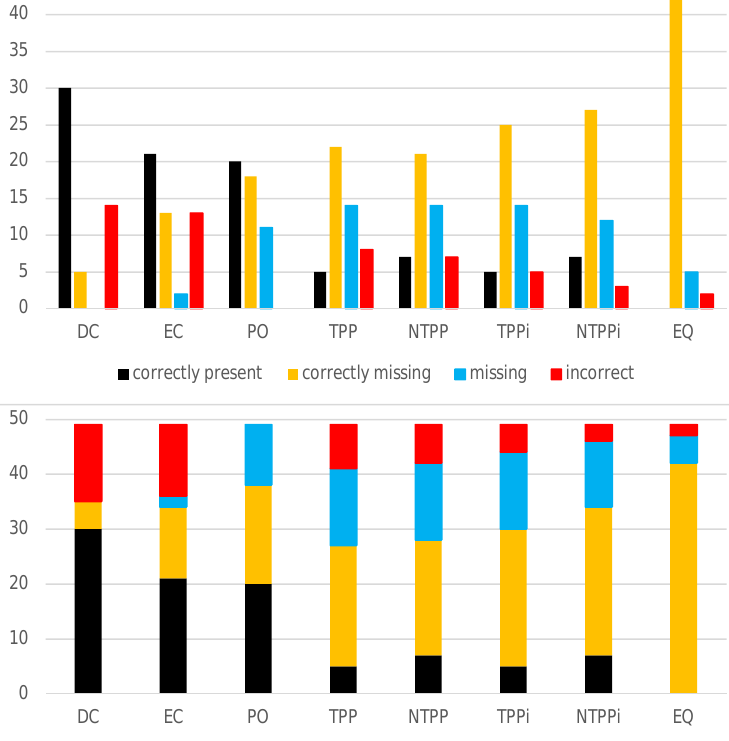} 
\caption{Relation statistics for the Composition Table for RCC-8 with anonymised relations produced by \chat. The upper chart shows the absolute number of relations, and the lower the relative percentage for each relation. 
 }
\label{xrcc8-comp-relation-stats}.
\end{figure}

In order to test whether the result was influenced by prior knowledge of RCC-8 gained as part of its training, we also performed the same experiment, but with all the relation names prefixed by an X to disguise the connection to RCC-8. The prompt was the same as above except for the change of relation names and the omission of the second sentence. The results are given in Fig. \ref{xrcc8-composition-table} while Fig. \ref{xrcc8-comp-relation-stats}  shows the absolute number and relative percentages of the 8 relations in the proportion that they are correctly or incorrectly predicted. As can be seen, \DC\  again is never missed as a correct prediction, and \EC\  is only missed twice; again \PO\  is never incorrectly predicted as present. As before, \DC, \EC, and  \PO   have the highest accuracies, along with \EQ, but \EQ\ is never predicted as present correctly, only incorrectly.  The overall average of correctly predicted relations (present and missing) drops from 71.94\% in the non-anonymous case above  to 67.09\% so there is some loss of performance though whether is due to the anonymisation of the relations or the stochasticity of \chat\  is not clear.

\subsection{Experiment \expnum: Preferred Compositions   in RCC-8}
As noted above, in general a composition of two relations will yield more than one possible base relations, but it turns out that humans tend to have a ``preferred'' relation.  For example, Ragni et al \shortcite{ragni2007cross} report on experiments performed on native German speakers and native Mongolian speakers for RCC-8. In their experiments the relations were described, but the human subjects were not allowed to draw possible configurations, so the setting is essentially equivalent to an LLM setting.

Given that humans may struggle to see all the possible relations\footnote{The fact that some humans may struggle to compute the composition table does not stop  it being a valid question to see if an LLM can determine the correct entries.}, determining whether there is agreement about the most preferred is good question to ask.  It turns out that there is good agreement in general across and within the two cultures, with the the percentage of people agreeing with the same preferred relation ranging from 30\% to 87.5\% (a random choice would yield 12.5\% on average since there are eight relations to choose from). (They did not query cases where the composition yields a unique relation, nor did they   consider \EQ\ as one of the two relations as this should be a trivial task.) This agreement is perhaps surprising since the two languages are linguistically very different. Ragni et al \shortcite{ragni2007cross} do report some differences though -- for example although both language speakers preferred \DC\  whenever it was consistent, Mongolians preferred \PO\  over \NTPPi\  whereas for Germans the converse was true. Both cultures only chose \EQ\  when composing a relation with its inverse (e.g. \TPP\  with \TPPi). 

The theory of \emph{preferred mental models} \cite{knauff1995preferred} states that people construct the simplest (computationally cheapest) model consistent with the premises. Their experiments showed that humans prefer models with the smallest overlapping complexity which explains the preference for \DC\  noted above.

Given the difficulty reported in Experiment 1 in \chat\  correctly inferring all possible relations in a composition, asking this more specific query seems a natural task to try; this gives some measure of whether the cognitive preferences of humans correspond at all to the choices made by \chat.  

The initial prompt given to \chat\   was identical to the initial prompt above when querying about composition. In its responses, sometimes \chat\  gave a clear preference for one relation, for example: 
\begin{quote}
   {\bf Prompt:} \emph{If DC(x,y) and DC(y,z) then what is  your preferred relationship between x and z?}
{\bf Response:} \emph{
The relations DC(x,y) and DC(y,z) mean that region x is disconnected from region y and region y is disconnected from region z. These relationships don't provide any information about the spatial relationship between x and z. x and z could be disconnected, they could touch at a boundary, or they could partially overlap, among other possibilities.
However, in absence of any other information and considering spatial continuity, I would say the most likely relation is DC(x, z), assuming x and z are also disconnected. Please note that this is just a statistical assumption; it's possible for x and z to have any of the other relations defined in RCC-8.}
\end{quote}
In other cases \chat\  justified its choice by saying it was the ``cautious'' choice, or the ``safest choice''. The humans in the \cite{ragni2007cross} experiment were not asked to justify their choices, but the claim made by the authors of that paper noted above is that choices were based on computational/simplicity considerations -- which differs from the reasons claimed here by \chat.

The initial prompt to \chat\  given was:
\begin{quote}
    \textit{Consider the following set of eight pairwise disjoint and mutually exhaustive binary spatial relations, between two one piece spatial 2D regions. These relations form part of the well known RCC-8 qualitative spatial reasoning calculus. DC(x,y) means that x and y are disconnected and share no spatial parts. EC(x,y) means that x and y touch at a boundary but do not share any interior parts. PO(x,y) means that x and y share a spatial part, but neither is part of the other.  TPP(x,y) means that x is part of y and touches y’s boundary. NTPP(x,y) means that x is part of y but does not touch y’s boundary.  TPPi (x,y) is the same as TPP(y,x).  NTPPi(x,y) is the same as NTPP(y,x). Finally, EQ(x,y) means that x and y are coincident. I will now ask you a series of questions about these relations. It is possible that in some cases there could be more than one relation that might hold;  in these cases just give me your single most preferred relationship. }
\end{quote}

\begin{figure}[htb]
\centering
\includegraphics[width=0.99\columnwidth]{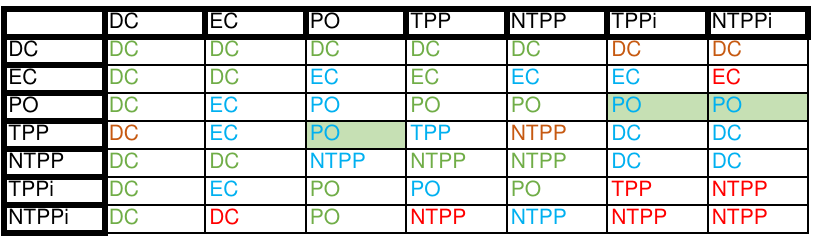} 
\caption{The table of preferred compositions for RCC-8 produced by \chat.
Green text means it agreed with the human most likely model. 			
Orange text means it agreed with the human most likely model but failed to notify that it was the only possible model. 			
Red means it chose an impossible relation.			
Blue means it chose a possible but not preferred model.		A green background means that it agreed with preferred relation of one of the language groups but not the overall average.		}
\label{rcc8preferred}.
\end{figure}

As can be seen in Fig. \ref{rcc8preferred}, \chat\ only agreed with the average human on 20/49 (40.82\%) though in a further three cases it agreed with one of the language groups (twice Mongolian, once German) giving a total of 23/49 (46.93\%). In seven cases it actually predicted an impossible relation as its preferred composition\footnote{Sometimes the humans in the \cite{ragni2007cross} experiment also predicted an impossible relation as can be seen in Table 2 of their publication.}, and in all cases where the composition is unique it failed to note that. In all the other cases, 18/49 (36.63\%) its preferred relation was possible but not preferred by either language group or overall. 

As can be seen, sometimes \chat\  made a logically inconsistent prediction, for example with \EC/\NTPPi; looking at the \chat's response in this case we can see that this is because it seems to misunderstand the meaning of \NTPPi:  
\begin{quote}
    \emph{If EC(x,y) and NTPPi(y,z), this means that region x touches the boundary of region y but does not overlap with it, and region y is a part of region z but does not touch z's boundary, with the perspective of y being contained in z.
In this scenario, it's likely that x and z are either disconnected (DC) or touch at the boundary (EC). Since y, which is part of z but does not touch z's boundary, is touching x at the boundary, there's a possibility for indirect spatial contact between x and z.
Without additional context, it's safer to assume that x and z touch at the boundary (EC). However, this is just an assumption, and the actual relationship between x and z could be different depending on the specific spatial arrangement.}
\end{quote}
In a different case, \TPPi/\TPPi, the initial understanding of the relations is correct inferring that $z$ is part of $x$, but then \chat\ makes a wrong step in the final part of its reasoning, and claims that \TPP($x,y)$:
\begin{quote}
    \emph{
If TPPi(x,y) and TPPi(y,z), this means that region y is a part of region x and touches x's boundary, and region z is a part of region y and touches y's boundary.
Given these conditions, it's likely that z is a part of x and touches x's boundary (TPP). This is because z, which is a part of y and touches y's boundary, is also indirectly a part of x through y, which is a part of x and touches x's boundary.
Therefore, my preferred relationship between x and z would be TPP(x,z), assuming that z is a part of x and touches x's boundary. Please note that this is an assumption and the actual relationship between x and z could be different depending on the specific spatial arrangement.}
\end{quote}
\chat\   makes exactly the same error with \TPPi/\NTPPi. For the wrong \NTPPi/\EC\  case, the
the description of its reasoning contains a more fundamental error:  
\begin{quote}
    \emph{...This is because y, which is a part of x and does not touch x's boundary, is also only touching z's boundary, so x and z would not have any overlap...}
\end{quote}
A similar more fundamental error is made in the cases of \pr{NTPPi} / \TPP, whilst in the final two wrong cases, \chat\  misunderstands the second relation as its inverse.

\begin{figure}[htb]
\centering
\includegraphics[width=0.99\columnwidth]{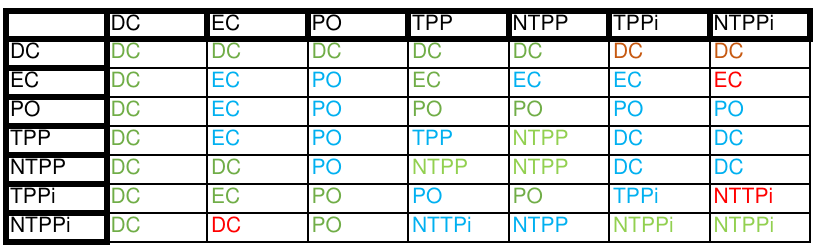} 
\caption{The table of preferred compositions for RCC-8 produced by \chat\   with anonymised relations. The relation names are coloured in the same way as in Table \ref{rcc8preferred}.
}
\label{xrcc8preferred}.
\end{figure}

In order to test whether the result was influenced by prior knowledge of RCC-8 gained as part of its training, we also performed the same experiment, but with all the relation names prefixed by an X to disguise the connection to RCC-8. The prompt was the same as above except for the change of relation names and the omission of the second sentence. The results are given in Fig. \ref{xrcc8preferred}.
\chat\ only agreed with the average human on 24/49 (48.99\%) though in a further five cases it agreed with one of the language groups (twice times Mongolian, twice German), and in one case the Mongolian preference was equal between the overall choice and the choice \chat\  made giving a an agreement with the German speakers of 26/49 (53.06\%) and 27/49 (55.1\%) with the Mongolian speakers.  In three cases (6.12\%) it actually predicted an impossible relation as its preferred composition (this also happened sometimes in the human experiment), and in two cases (out of 11) where t    he composition is unique it failed to note that. In all the other cases, 20/49 (60.82\%) its preferred relation was possible but not preferred by either language group or overall.   The overall agreement with humans in this anonymised case is actually significantly higher than the non-anonymised experiment above; it is not clear if there is a reason for this and if so what.

\section{Experiment \expnum: Spatial Continuity}
Continuity networks were introduced by Randell and Cohn\shortcite{randell1989modelling} to represent the set of possible `next’ relations that might obtain, assuming that motion is continuous and any transformations in the shape and/or size of an object are also continuous.  Subsequently these have been termed ``conceptual neighbourhoods’’ \cite{freksa1992temporal}.  Fortunately, the abbreviation for both of these terms is CN, which we shall use henceforth.  The CN for RCC-8 is depicted in Fig. \ref{rcc8-cn-diagram}.  The initial prompt was the same as for the composition table prompt above, but with two additional sentences at the end: \emph{ In all these questions, assume that all motion is continuous, and any deformations of the spatial regions are also continuous. Also please describe the motion  or changes of size or shape that would be required for the change of relation to take place}. Each subsequent prompt also emphasises the need for the relation to be an immediate next relation.   As an example prompt/response pair consider the \EC\ case:

\begin{quote}
    {\bf Prompt:} \textit{If EC(a,b) holds, then what are the next possible relations that might hold as the two regions translate or change their shape or size?  I am interested in the immediate next relation, so no intermediate relation must hold.}
{\bf Response:}
\textit{If EC(a,b) holds, which means that the two regions a and b are just touching at a boundary but do not share any interior parts, then the immediate next possible relations as the two regions translate or change their shape or size could be either DC(a,b) or PO(a,b).
DC(a,b): This would occur if the regions move away from each other and are no longer in contact at all, becoming fully disconnected.
PO(a,b): This would occur if one region begins to overlap with the other, but neither region is completely contained within the other. They would then be partially overlapping, sharing some but not all of their spatial parts.
The transitions to these relations would involve either moving the regions apart or moving them further into each other. In the case of transition to PO(a,b), there could also be some deformation of the regions if they change shape to allow for the partial overlap.}
\end{quote}
An example of a faulty prediction is from \PO:
\begin{quote}
    \emph{If PO(a,b) holds, indicating that regions a and b partially overlap but neither is fully contained within the other, the immediate next relations could be EC(a,b), DC(a,b), TPP(a,b), TPPi(a,b), or EQ(a,b).
EC(a,b): This would occur if the overlapping parts of a and b are reduced to just boundary parts, meaning the regions are touching but no longer overlapping.
DC(a,b): This would occur if regions a and b are moved or deformed such that they no longer overlap or touch at all.
TPP(a,b) or TPPi(a,b): This would occur if one region becomes fully contained within the other, either by moving into it or shrinking. TPP(a,b) means a is part of b and touches b’s boundary. TPPi(a,b) means b is part of a and touches a’s boundary.
EQ(a,b): This could occur if the regions are reshaped or resized such that they coincide exactly, becoming the same region in space.
The transitions to these relations would involve either moving the regions, resizing them, or reshaping them in some way.}
\end{quote}
Here, \chat\  has mistakenly  said that \DC\  is a next relation, failing to notice that it has to pass through \EC\  on the way.

As can be seen, the predictions are mostly correct with just five errors: three where a link is falsely predicted and two where a link is missing. The two missing links are both from \EQ\  (to \NTPP\    and to \NTPPi).  Interestingly the links in the reverse direction are correctly predicted, so \chat\  is not able to reason that a link in one direction implies the link in the other direction should also  be present. Similarly whilst the absence of a link from \DC\  to \PO\  is correctly predicted, the inverse case is not, and is also the case for \PO\  to/from \NTPP\  and \NTPPi.


\begin{figure}[htb]
\centering
\includegraphics[width=0.99\columnwidth]{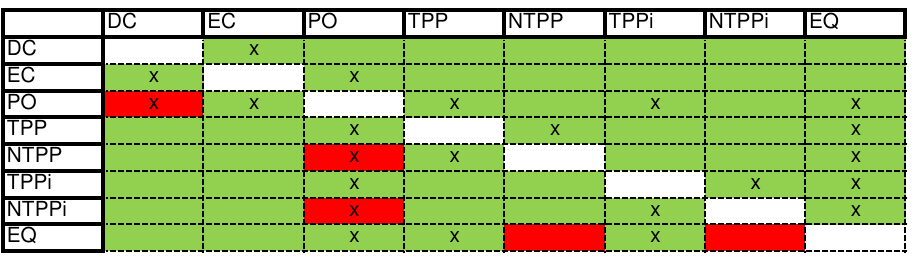} 
\caption{The Continuity Table for RCC8 produced by \chat. An `x' means that the  relation in that column is predicted as an immediate neighbour of the relation in that row.  An empty box means that the relation is not predicted as an immediate neighbour.  Green means that the prediction was correct and red that it was incorrect. The leading diagonal is white since a relation is not a next relation of itself.}
\label{rcc8CN}.
\end{figure}

\begin{figure}[htb]
\centering
\includegraphics[width=0.99\columnwidth]{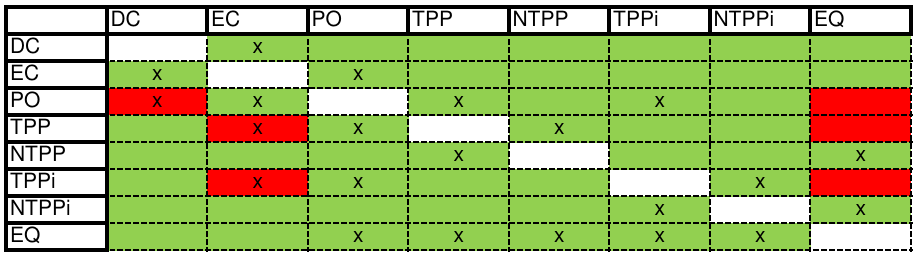} 
\caption{The Continuity Table for RCC-8 produced by \chat\  using disguised relation names.  The meaning of the colouring is the same as in Fig. \ref{rcc8CN}.
}
\label{xrcc8CN}.
\end{figure}

In order to test whether the result was influenced by prior knowledge of RCC-8 gained as part of its training, we also performed the same experiment, but with all the relation names prefixed by an X to disguise the connection to RCC-8. The prompt was the same as above except for the change of relation names and the omission of the second sentence. The results are given in Fig. \ref{xrcc8CN}. There are 3 incorrectly predicted links, 3 missing links, 
19 correctly predicted links and 
31 correct missing links, giving an accuracy of 50/56 (89.2\%).  This is slightly worse than the case above. There is one more missing link but the missing links are all different in the two cases. Although there are the same number of wrong links, only one of these is in common (\PO\  to \DC). Overall the results are broadly similar and may be due to the stochastic nature of \chat's responses, suggesting that either the disguise was not very effective, or that prior training did not really affect the response and it was able to reason from `first principles' (if not always correctly) in response to each prompt.

\section{Concluding Remarks and Future Work}

This investigation has supported the widely-held view that LLMs can struggle to do reasoning tasks\footnote{ Bender et al \shortcite{bender2021dangers} have observed that LLMs might be regarded just as ``stochastic parrots'' and thus it is not suprising that precise, logically correct deductive reasoning is challenging for an LLM.}. In the case of Experiment 1, in which \chat\  was asked to compute the entire composition table for RCC-8, this is a non trivial task even for humans, so it is perhaps not surprising that \chat\  did not achieve 100\% accuracy -- the scores of 71.94\% (and 67.09\% for the anonymised relatins) are clearly much better than chance and do suggest a reasonable facility to perform such computations.   A detailed analysis of the actual conversations in the supplementary material shows that sometimes \chat\  does appear able to do some interesting (qualitative) spatial reasoning, but often fails, sometimes making elementary mistakes.  It also shows inconsistency in being able to reason correctly about a relation but not its inverse.  It also sometimes confuses a relation with its inverse.   It is possible that fine tuning, explicit chain-of-thought prompting, or more carefully engineered prompts might improve performance; however, given the stochastic nature of LLMs it seems unlikely  that the results would be as good as logical reasoning (the experiment on preferred relations is of course not strictly a logical reasoning exercise, except for the requirement not to predict spatially impossible relations).

There are a variety of avenues for further work which present themselves. Other calculi could be experimented with  -- for example the coarser calculus RCC-5, or calculi for reasoning about direction or size \cite{cohn2008qualitative}. Other LLMs could be evaluated -- though since  new LLMs and new LLM versions are continually being released, this is a challenge with no definite stopping point. Tracking the change in performance of a particular LLM across releases would also be of interest -- though in the case of closed LLMs such as \chat\  where the owners have the right to harvest user conversations and use them for future training, it will not be clear if any improvement is the result of leakage from the previous conversation or more general performance improvement\footnote{However, note that no feedback was given to \chat\  as to whether the proffered response was correct or not.}.  It has already been observed \cite{cohn2023dialectical} that different LLMs have different strengths -- determining which LLMs are better at which spatial reasoning tasks would also be worth of future investigation. The overall conclusion that LLMs in general struggle with more complex spatial reasoning tasks is likely to remain the case, at least for the foreseeable future. In the API version of GPT, different temperatures could be tried, and multiple runs with averages computed. Different prompts and prompting strategies could be tried, though arguably since QSR has always been viewed as a form of commonsense reasoning, it should not be necessary to devise specific prompts to elicit commonsense behaviour. 

It is not clear how successful the anonymisation was -- in one case I mistyped an X relation and it was able to suggest the intended relation name, suggesting that it has the ability to dissect relation names; thus more sophisticated anonymisation might be tried. In earlier work \cite{cohn2023dialectical} we had already done some limited experimentation asking an LLM to reason about spatial relations in a real world context rather than the purely abstract setting used in the experiments in this paper -- it would be interesting to conduct more extensive tests  LLMs doing compositional reasoning in a more realistic setting, and similarly for the continuity experiment.

 Experiment 2 above  already investigated how LLM performance compared to human performance to a limited extent but further investigation would be worthwhile, including a head-to-head comparison rather than simply taking a result from the literature originally intended to investigate a different question.  Another interesting avenue for further work will be to explore the use of multimodal FMs -- when humans perform spatial reasoning tasks including the challenge of building a composition table, it is natural to use pencil and paper to sketch diagrams and possible scenarios -- investigating whether a multi-modal FM with such abilities (including the ability to analyse its own drawings) would be of great interest to the spatial reasoning community.

 As mentioned above, another possible avenue of research is to investigate different prompting strategies, including k-shot \cite{dang2022prompt}, chain-of-thought\cite{wei2022chain} and tree-of-thought\cite{yao2023tree} strategies. Not doing so was deliberate in this paper as I was interested in exploring in how the ``vanilla'' LLM would perform. Whilst for specific downstream tasks, fine-tuning or employing specific prompting strategies may reasonable, there is an argument to be made that for commonsense reasoning, this is not a reasonable strategy since the task is a general one rather than a specific downstream task.

\section{Data statement}
All the conversations with \chat\  that support the summary tables in this paper can be found at http://tinyurl.com/qr23sup.

\section{Acknowledgments}
This work was supported by: The Alan Turing Institute; the   Economic and Social Research Council (ESRC) under grant ES/W003473/1; the Turing’s Defence and Security programme through a partnership with the UK government in accordance with the framework agreement between GCHQ and  The Alan Turing Institute.

\bibliography{qr}

\end{document}